\pdfoutput=1
\documentclass[11pt,letterpaper]{article}

\usepackage[top=2.5cm,bottom=2.5cm,left=2.6cm,right=2.6cm]{geometry}
\usepackage{times}
\usepackage{microtype}
\usepackage{amsmath,amssymb,amsfonts,amsthm}
\usepackage{graphicx}
\usepackage{booktabs}
\usepackage{multirow}
\usepackage{array}
\usepackage{url}
\usepackage{xcolor}
\usepackage[pagebackref=false,breaklinks=true,colorlinks,bookmarks=false,citecolor=blue!70!black,linkcolor=red!70!black,urlcolor=blue!70!black]{hyperref}
\usepackage{caption}
\usepackage{subcaption}
\usepackage{enumitem}
\usepackage{tikz}
\usepackage{algorithm}
\usepackage{algpseudocode}
\usepackage{fancyhdr}
\usepackage{cite}

\newtheorem{theorem}{Theorem}

\newtheorem{proposition}[theorem]{Proposition}
\newtheorem{definition}{Definition}

\fancypagestyle{plain}{%
  \fancyhf{}%
  \fancyfoot[C]{\thepage}%
}

\begin{document}

\title{\textbf{\LARGE loopxDINO: Dense Visual Prediction via Iterative Latent Reasoning}}

\author{
  \textbf{Xiangjing An}\thanks{Corresponding author: \texttt{axjing.ai@gmail.com}}\qquad
  \textbf{Zhibin Jiao}\qquad
  \textbf{Aiting Chen}\\[0.4em]
  {\small \texttt{\{axjing.ai@gmail.com, jzb1631@163.com, ChengAiTiny@163.com\}}}
}

\date{}

\maketitle

\begin{abstract}
Standard dense visual prediction architectures typically scale representation capacity by stacking deep feedforward decoders (often 6 to 16 transformer layers comprising 50M--200M parameters). However, this explicit layer stacking incurs substantial activation memory overhead during backpropagation, demands high off-chip memory bandwidth to fetch non-tied weights from DRAM, and imposes a static computational allocation that fails to adapt to spatial image complexity.
In this work, we present \textbf{loopxDINO}, an architecture that substitutes physical decoder depth with \textbf{iterative latent reasoning}. Visual computation is structured into two decoupled components: (1) a frozen foundation backbone (DINOv2-Base) that extracts spatial features in a single forward pass and caches static Key-Value representations in GPU memory; and (2) a compact, weight-tied \textbf{Looped Cross-Attention Decoder} (7.96M core parameters, 14.85M total trainable parameters) that recurrently updates learned queries across $T$ iterations.
Recurrence dynamics are stabilized through step-conditioned Adaptive Layer Normalization (adaLN), LayerNorm-induced hyper-ellipsoid manifold regularization, and learnable residual contraction damping. Intermediate mask predictions are routed back as additive spatial attention biases in subsequent cross-attention steps, directing queries toward ambiguous object contours without introducing additional parameters.
On the Pascal VOC 2012 semantic segmentation benchmark under single-GPU hardware constraints (NVIDIA GeForce RTX 5060 Ti 16GB), loopxDINO ($T=2$) achieves \textbf{85.66\% mIoU} and \textbf{65.01\% Boundary IoU}. This exceeds a 6-layer stacked transformer baseline (\textbf{DINO-Trans6L}, 54.06M trainable parameters) by \textbf{+1.09\% mIoU} and \textbf{+7.41\% Boundary IoU} with a \textbf{72.5\% parameter reduction}. Relative to a 4-layer convolutional spatial probe (\textbf{DINO-FCN4L}, 2.43M parameters, 83.82\% mIoU) that calibrates the linear separability of the frozen backbone, loopxDINO substantially enhances boundary delineation (+6.43\% bIoU) and occlusion resilience (+6.76\% mIoU under 60\% patch masking). Equipped with Adaptive Computation Time (ACT), loopxDINO achieves \textbf{87.3 FPS} and maintains \textbf{29.64\% mIoU} under 60\% occlusion (+5.83\% over DINO-Trans6L).
The complete source code, pre-trained weights, and reproducible evaluation suites are open-sourced under the Apache~2.0 license at \href{https://github.com/loopxDINO/loopxDINO}{\url{https://github.com/loopxDINO/loopxDINO}}.
\end{abstract}

\section{Introduction}
\label{sec:intro}

Dense visual perception tasks, including semantic segmentation and boundary delineation, commonly employ deep feedforward decoders atop pre-trained vision backbones~\cite{dosovitskiy2020image,oquab2023dinov2,kirillov2023segment,cheng2022masked}. While scaling physical layer depth expands representational breadth, it introduces two practical system bottlenecks:
\begin{enumerate}[leftmargin=*]
    \item \textbf{Memory and Bandwidth Overhead:} Non-tied weight matrices must be transferred from off-chip DRAM at each successive layer. In addition, intermediate forward activations accumulate linearly during backpropagation, creating memory bottlenecks on commodity hardware (e.g., 16\,GB VRAM).
    \item \textbf{Rigid Computational Allocation:} Standard feedforward decoders dedicate an invariant computational budget per token regardless of local visual ambiguity~\cite{graves2016adaptive,dehghani2018universal}; uniform background patches receive identical computational depth to intricate boundaries between overlapping objects.
\end{enumerate}

Recurrent processing offers an alternative to physical depth expansion by allowing a parameter-shared layer to iteratively refine latent representations across dynamic computational horizons~\cite{dehghani2018universal,giannou2023looped}. Motivated by test-time computation scaling in reasoning and sequence models~\cite{graves2016adaptive,deepseek2025r1,openai2024o1,sun2024learning}, we investigate whether computational depth in dense visual prediction can be achieved through recurrent latent iteration rather than explicit layer stacking.

\begin{figure}[htbp]
  \centering
  \includegraphics[width=\linewidth]{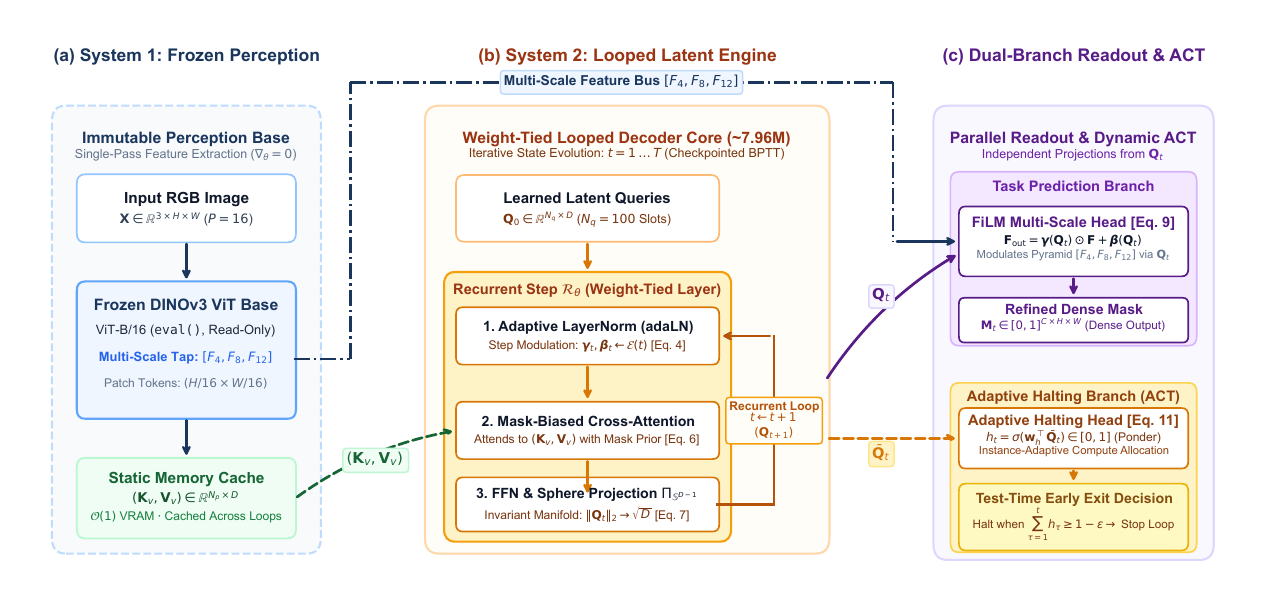}
  \caption{\textbf{Architectural overview of loopxDINO.} \textbf{(a) Immutable Perception Base:} An input image $\mathbf{X} \in \mathbb{R}^{3 \times H \times W}$ is processed in a single forward pass by a frozen foundation backbone ($\nabla_{\theta}=0$), producing spatial features and a persistent Key-Value cache $(\mathbf{K}_v, \mathbf{V}_v) \in \mathbb{R}^{H_p W_p \times D}$. \textbf{(b) Looped Latent Reasoning Engine:} A compact, weight-tied Looped Cross-Attention Decoder ($\approx 7.96\,\text{M}$ core parameters) iteratively refines learned queries $\mathbf{Q}_t$ across $t=1 \dots T$. Recurrent updates are modulated by step-conditioned Adaptive Layer Normalization ($\mathrm{adaLN}$), regularized by LayerNorm hyper-ellipsoid projection, and guided by spatial attention biases derived from intermediate mask predictions. \textbf{(c) Readout \& Dynamic ACT:} Multi-scale representations from layers 4, 8, and 12 are fused to generate dense masks $\mathbf{M}_t$, while a halting controller predicts termination probabilities $h_t$ for adaptive computational allocation.}
  \label{fig:architecture}
\end{figure}

We introduce \textbf{loopxDINO}, an iterative latent reasoning architecture for dense visual prediction. Our main contributions are:
\begin{enumerate}[leftmargin=*]
    \item \textbf{Decoupled Foundation Perception with Static KV-Cache Amortization:} We separate frozen feature extraction from iterative query refinement, projecting spatial Key-Value representations once into fast SRAM rather than recomputing them per layer as in standard decoders. Without this caching, recurrent attention requires redundant linear projections scaling with $\mathcal{O}(T \cdot N_p D)$; our formulation amortizes bandwidth to $\mathcal{O}(N_p D)$ with identical expressivity.
    \item \textbf{Looped Cross-Attention with Closed-Loop Spatial Feedback:} We replace feedforward transformer stacks with a single weight-tied cross-attention layer ($\approx 7.96\,\text{M}$ core parameters) stabilized via step-conditioned adaLN, LayerNorm hyper-ellipsoid regularization, and residual contraction damping. Unlike unconstrained recurrence that quickly degenerates into over-smoothing, this formulation guarantees bounded state velocity while repurposing intermediate mask logits as parameter-free spatial attention biases that actively preserve feature diversity.
    \item \textbf{Empirical Characterization of Compute Depth and Test-Time Scaling:} We demonstrate on Pascal VOC 2012 that parameter depth can be effectively substituted with compute depth: with 72.5\% fewer trainable parameters than a 6-layer stacked baseline (DINO-Trans6L), loopxDINO ($T=2$) achieves 85.66\% mIoU (+1.09\%) and 65.01\% Boundary IoU (+7.41\%). Furthermore, while clean images saturate at $T=2$ due to intrinsic feature separability, severe 60\% occlusion reveals the necessity of deeper test-time computation, where $T=4$ achieves 29.64\% mIoU (+9.87\% over $T=2$ and +5.83\% over DINO-Trans6L). Practical deployment is supported by activation checkpointing ($\mathcal{O}(1)$ activation memory) and static CUDA Graph execution (up to 408.2 FPS).
\end{enumerate}

\section{Related Work}
\label{sec:related}

\paragraph{Visual Foundation Models.}
Self-supervised vision transformers, pioneered by DINO~\cite{caron2021emerging}, MAE~\cite{he2022masked}, and DINOv2~\cite{oquab2023dinov2}, produce rich patch-level representations without manual supervision. DINO features exhibit emergent semantic segmentation properties and spatial consistency across natural scenes. However, downstream dense prediction tasks commonly deploy these features alongside heavy task-specific decoders (e.g., UPerNet, MaskFormer~\cite{cheng2021per}, Mask2Former~\cite{cheng2022masked}) containing 50M to 200M trainable parameters. loopxDINO shows that a 7.96M parameter-tied recurrent decoder adequately harnesses and refines these representations without feedforward stacking.

\paragraph{Recurrent \& Universal Transformers.}
Weight-tied recurrence in transformer architectures originated in sequence modeling with Universal Transformers~\cite{dehghani2018universal} and Perceiver architectures~\cite{jaegle2021perceiver,jaegle2021perceiverio}. Theoretical studies have demonstrated that looped transformers can emulate algorithmic procedures such as gradient descent and in-context learning~\cite{giannou2023looped,yang2023looped}. Concurrently, state space models~\cite{gu2023mamba,zhu2024vision} and test-time training (TTT)~\cite{sun2024learning} investigate recurrent state updates for linear-time sequence processing. loopxDINO adapts weight-tied recurrence to visual dense prediction, establishing that recurrent latent updates can match or exceed deep feedforward decoders in boundary resolution and parameter efficiency.

\paragraph{Over-Smoothing in Deep Architectures.}
In deep graph neural networks and deep transformers, over-smoothing represents a primary degradation mode~\cite{rusch2023survey}: repeated aggregation through unconstrained attention matrices behaves as a contractive low-pass filter, dissipating Dirichlet energy and collapsing token representations into a homogeneous subspace. Unrolled recurrent decoders face an analogous failure mode. loopxDINO addresses this challenge through step-conditioned adaLN modulation, LayerNorm hyper-ellipsoid regularization, and closed-loop spatial mask biases that preserve token variance.

\paragraph{Deep Equilibrium Models \& Fixed-Point Networks.}
Deep Equilibrium Models (DEQs)~\cite{bai2019deep,bai2020multiscale} eliminate depth by solving for the infinite-depth fixed point of a single layer using root-finding algorithms (e.g., Broyden's method). While conceptually unified, standard DEQs often encounter numerical instability, slow backward Jacobian evaluations, and high training latency. loopxDINO instead employs an unrolled, contractive recurrence with explicit loop-step conditioning (\textit{adaLN}), attaining rapid convergence within 2 to 4 steps while retaining exact gradient backpropagation.

\paragraph{Dynamic Test-Time Computation.}
Adaptive Computation Time (ACT)~\cite{graves2016adaptive}, alongside PonderNet~\cite{banino2021pondernet} and Depth-Adaptive Transformers~\cite{elbayad2019depth}, provides recurrent models with learned halting mechanisms. Concurrently, dynamic test-time computation has proven essential in recent reasoning models~\cite{deepseek2025r1,openai2024o1}. loopxDINO incorporates these principles into visual dense prediction, allowing visual queries to scale iteration depth dynamically on challenging image regions.

\section{Architecture and Method}
\label{sec:method}

\subsection{System 1: Immutable Perception Base}
System~1 performs high-capacity visual feature extraction from input image $\mathbf{X} \in \mathbb{R}^{3 \times H \times W}$. Rather than training the backbone end-to-end, loopxDINO employs a frozen pre-trained DINOv2-Base vision transformer $\mathcal{F}_{\theta_1}$ (ViT-Base/14, 86.58M parameters)~\cite{oquab2023dinov2}:
\begin{equation}
    \label{eq:backbone_feature}
    \mathbf{F} = \mathcal{F}_{\theta_1}(\mathbf{X}) \in \mathbb{R}^{B \times N_p \times D}, \quad \text{with } \nabla_{\theta_1} = 0,
\end{equation}
where $N_p = \frac{H}{P} \times \frac{W}{P}$ denotes the number of non-overlapping spatial patches of size $P \times P$ ($P=14$), and $D$ is the embedding dimension ($D=768$).
To accommodate foundation models that incorporate auxiliary tokens (such as the 4 or 8 register tokens in DINOv2), spatial tokens are isolated by taking the trailing $H_p \cdot W_p$ elements:
\begin{equation}
    \label{eq:spatial_feature}
    \mathbf{F}_{\text{spatial}} = \mathbf{F}[:, -(H_p \cdot W_p):, :],
\end{equation}
which strips preceding class and register tokens while maintaining spatial grid alignment across backbone checkpoints.

\paragraph{Static KV-Cache Amortization.}
Standard feedforward decoders recompute Key ($\mathbf{K}$) and Value ($\mathbf{V}$) projections at every transformer layer. Because spatial representation $\mathbf{F}_{\text{spatial}}$ in Eq.~(\ref{eq:spatial_feature}) remains static throughout recurrence, linear projections are evaluated once prior to entering the iterative loop:
\begin{equation}
    \label{eq:kv_projection}
    \mathbf{K}_v = \mathbf{F}_{\text{spatial}} \mathbf{W}_K^\top, \quad \mathbf{V}_v = \mathbf{F}_{\text{spatial}} \mathbf{W}_V^\top,
\end{equation}
where $\mathbf{W}_K, \mathbf{W}_V \in \mathbb{R}^{D \times D}$. These projections are cached in GPU memory, reducing aggregate cross-attention memory bandwidth across $T$ iterations from $\mathcal{O}(T \cdot N_p D)$ to $\mathcal{O}(N_p D)$.

\subsection{System 2: Looped Cross-Attention Decoder}
System~2 is realized as a parameter-shared cross-attention block $\mathcal{T}_{\theta_2}$ ($\approx 7.96\,\text{M}$ core parameters for $D=768$).
Given learned latent queries $\mathbf{Q}_0 \in \mathbb{R}^{B \times N_q \times D}$ ($N_q=21$ for Pascal VOC) initialized from $\mathcal{N}(0, 0.02)$, query representations are iteratively updated across $T$ cycles:
\begin{equation}
    \label{eq:looped_decoder}
    \mathbf{Q}_t = \mathcal{T}_{\theta_2}(\mathbf{Q}_{t-1}, \mathbf{K}_v, \mathbf{V}_v; t), \quad t \in \{1, 2, \dots, T\}.
\end{equation}

\subsubsection{Loop-Step Conditioning via Adaptive LayerNorm}
Directly executing identical weights on unmodulated query state $\mathbf{Q}_{t-1}$ promotes over-smoothing, causing queries to collapse toward an undifferentiated centroid~\cite{dehghani2018universal,rusch2023survey}. To differentiate representations across iterations, an \textbf{Adaptive Layer Normalization (adaLN)} mechanism modulates latent queries based on step index $t$. A continuous temporal embedding layer $\mathcal{E}_{\text{step}}(t)$ projects discrete step $t$ into six affine modulation vectors:
\begin{equation}
    \label{eq:step_embedding}
    [\boldsymbol{\gamma}_1^t, \boldsymbol{\beta}_1^t, \boldsymbol{\gamma}_2^t, \boldsymbol{\beta}_2^t, \boldsymbol{\gamma}_3^t, \boldsymbol{\beta}_3^t] = \mathcal{E}_{\text{step}}(t) \in \mathbb{R}^{6D}.
\end{equation}
Each sub-layer (Self-Attention, Cross-Attention, Feedforward Network) is modulated dynamically before execution:
\begin{align}
    \label{eq:adaln_sa}
    \widehat{\mathbf{Q}}_{t-1}^{(1)} &= \text{adaLN}(\mathbf{Q}_{t-1}; \boldsymbol{\gamma}_1^t, \boldsymbol{\beta}_1^t) = (1 + \boldsymbol{\gamma}_1^t) \odot \text{LN}(\mathbf{Q}_{t-1}) + \boldsymbol{\beta}_1^t, \\
    \label{eq:sa_update}
    \mathbf{Q}_{t-1}^{(\text{sa})} &= \mathbf{Q}_{t-1} + \text{SelfAttention}(\widehat{\mathbf{Q}}_{t-1}^{(1)}), \\
    \label{eq:adaln_ca}
    \widehat{\mathbf{Q}}_{t-1}^{(2)} &= \text{adaLN}(\mathbf{Q}_{t-1}^{(\text{sa})}; \boldsymbol{\gamma}_2^t, \boldsymbol{\beta}_2^t), \\
    \label{eq:ca_update}
    \mathbf{Q}_{t-1}^{(\text{ca})} &= \mathbf{Q}_{t-1}^{(\text{sa})} + \text{CrossAttention}(\widehat{\mathbf{Q}}_{t-1}^{(2)}, \mathbf{K}_v, \mathbf{V}_v), \\
    \label{eq:adaln_ffn}
    \widehat{\mathbf{Q}}_{t-1}^{(3)} &= \text{adaLN}(\mathbf{Q}_{t-1}^{(\text{ca})}; \boldsymbol{\gamma}_3^t, \boldsymbol{\beta}_3^t), \\
    \label{eq:ffn_damping}
    \mathbf{Q}_t^{\text{raw}} &= \alpha \cdot \left(\mathbf{Q}_{t-1}^{(\text{ca})} + \text{FFN}(\widehat{\mathbf{Q}}_{t-1}^{(3)})\right) + (1 - \alpha) \cdot \mathbf{Q}_{t-1}^{(\text{ca})},
\end{align}
where $\alpha \in [0, 1]$ in Eq.~(\ref{eq:ffn_damping}) serves as a learnable residual contraction parameter that bounds the Lipschitz constant of the recurrent mapping.

\subsubsection{Hyper-Ellipsoid Manifold Regularization via LayerNorm}
To maintain bounded activations across arbitrary unroll horizons $T$, each iteration concludes with LayerNorm regularization:
\begin{equation}
    \label{eq:hypersphere_norm}
    \mathbf{Q}_t = \text{LayerNorm}(\mathbf{Q}_t^{\text{raw}}).
\end{equation}
LayerNorm standardizes each token vector $\mathbf{q} \in \mathbb{R}^D$ to zero mean and unit variance: $\bar{\mathbf{q}} = \frac{\mathbf{q} - \boldsymbol{\mu}}{\sqrt{\boldsymbol{\sigma}^2 + \epsilon}}$, restricting $\bar{\mathbf{q}}$ to the intersection of hyperplane $\mathbf{1}^\top \bar{\mathbf{q}} = 0$ and the sphere of radius $\sqrt{D}$. With learnable scale $\boldsymbol{\gamma}_{\text{LN}} \in \mathbb{R}^D$ and shift $\boldsymbol{\beta}_{\text{LN}} \in \mathbb{R}^D$, normalized output $\mathbf{z} = \bar{\mathbf{q}} \odot \boldsymbol{\gamma}_{\text{LN}} + \boldsymbol{\beta}_{\text{LN}}$ is confined to an affine compact \textbf{hyper-ellipsoid manifold}:
\begin{equation}
    \label{eq:hyper_ellipsoid_manifold}
    \mathcal{E}_D = \left\{ \mathbf{z} \in \mathbb{R}^D \;\middle|\; \sum_{j=1}^D \frac{(z_j - \beta_{\text{LN}, j})^2}{\gamma_{\text{LN}, j}^2} = D, \quad \sum_{j=1}^D \frac{z_j - \beta_{\text{LN}, j}}{\gamma_{\text{LN}, j}} = 0 \right\}.
\end{equation}
As established in Section~\ref{sec:theory}, Eq.~(\ref{eq:hyper_ellipsoid_manifold}) confines the sequence $\{\mathbf{Q}_t\}_{t=1}^T$ to an invariant domain $\|\mathbf{Q}_t\|_F \le \sqrt{N_q D} \|\boldsymbol{\gamma}_{\text{LN}}\|_\infty + \sqrt{N_q} \|\boldsymbol{\beta}_{\text{LN}}\|_2$, precluding activation divergence.

\subsection{Multi-Scale Pyramid Fusion and Mask-Guided Attention}
\label{subsec:multiscale_mask_guidance}

\paragraph{Multi-Scale Feature Pyramid Fusion.}
Although the final layer ($L=12$) of the frozen backbone ($\nabla_{\theta_1} = 0$) supplies rich semantic abstractions, its spatial resolution is comparatively coarse ($37 \times 37$ patch grid for $512 \times 512$ inputs at $P=14$). To recover high-frequency boundary geometry, loopxDINO extracts intermediate representations from shallow, intermediate, and deep layers $l \in \{4, 8, 12\}$:
\begin{equation}
    \label{eq:multiscale_features}
    \mathbf{F}_4, \mathbf{F}_8, \mathbf{F}_{12} = \mathcal{F}_{\theta_1}^{(4, 8, 12)}(\mathbf{X}), \quad \mathbf{F}_l \in \mathbb{R}^{B \times N_p \times D}.
\end{equation}
Lateral $1 \times 1$ convolutions project these feature maps ($D=768$) to a unified dimension $d_{\text{feat}}=256$ before merging through top-down addition:
\begin{align}
    \label{eq:fpn_p12}
    \mathbf{P}_{12} &= \text{Conv}_{1\times 1}(\mathbf{F}_{12}), \\
    \label{eq:fpn_p8}
    \mathbf{P}_8 &= \text{Conv}_{1\times 1}(\mathbf{F}_8) + \mathbf{P}_{12}, \\
    \label{eq:fpn_p4}
    \mathbf{P}_4 &= \text{Conv}_{1\times 1}(\mathbf{F}_4) + \mathbf{P}_8, \\
    \label{eq:fpn_fused}
    \mathbf{F}_{\text{fused}} &= \text{Conv}_{3\times 3}(\mathbf{P}_4),
\end{align}
followed by sub-pixel convolution upsampling that restores fine spatial resolution with modest parameter cost (4.46M parameters).

\paragraph{Mask-Guided Cross-Attention Bias.}
At each recurrence step $t$, intermediate query states generate dense mask predictions $\mathbf{M}_t \in \mathbb{R}^{B \times N_q \times H \times W}$. Rather than discarding these predictions, loopxDINO establishes a recurrent feedback connection by converting intermediate masks into additive spatial attention biases for the subsequent iteration:
\begin{align}
    \label{eq:mask_softmax}
    \mathbf{S}_t &= \text{Softmax}(\mathbf{M}_t, \text{dim}=\text{class}), \\
    \label{eq:attention_bias}
    \mathbf{B}_{t+1} &= \lambda_{\text{bias}} \cdot \left[ 0.5 \cdot \text{AvgPool}(\mathbf{S}_t) + 0.5 \cdot \text{MaxPool}(\mathbf{S}_t) \right],
\end{align}
where combined average and max pooling downsamples the spatial resolution to the patch grid $(H_p, W_p)$ while preserving thin structures. Given key dimension $d_k = D / N_{\text{heads}}$ ($d_k = 768 / 12 = 64$), the resulting tensor $\mathbf{B}_{t+1} \in \mathbb{R}^{B \times N_{\text{heads}} \times N_q \times N_p}$ is incorporated directly into native scaled dot-product cross-attention:
\begin{equation}
    \label{eq:guided_attention}
    \text{Attn}(\mathbf{Q}, \mathbf{K}_v, \mathbf{V}_v) = \text{Softmax}\left(\frac{\mathbf{Q} \mathbf{K}_v^\top}{\sqrt{d_k}} + \mathbf{B}_{t+1}\right) \mathbf{V}_v,
\end{equation}
which steers semantic queries toward ambiguous boundary zones without introducing additional decoder parameters.

\paragraph{Multi-Step Hybrid Cascade Loss.}
To maintain stable gradient propagation across unrolled iterations, intermediate step predictions are supervised via a combined Cross-Entropy and Multi-Class Dice cascade loss:
\begin{equation}
    \label{eq:step_loss}
    \mathcal{L}_{\text{step}}(\mathbf{M}_t, \mathbf{Y}) = \mathcal{L}_{\text{CE}}(\mathbf{M}_t, \mathbf{Y}) + \lambda_{\text{dice}} \mathcal{L}_{\text{Dice}}(\mathbf{M}_t, \mathbf{Y}),
\end{equation}
\begin{equation}
    \label{eq:cascade_loss}
    \mathcal{L}_{\text{cascade}} = \sum_{t=1}^T w_t \cdot \mathcal{L}_{\text{step}}(\mathbf{M}_t, \mathbf{Y}), \quad \text{with } w_t = \frac{2t}{T(T+1)},
\end{equation}
where weighting factor $w_t$ in Eq.~(\ref{eq:cascade_loss}) increases linearly with step index, focusing parameter optimization progressively toward late-stage representation refinement.

\begin{figure}[htbp]
  \centering
  \includegraphics[width=\linewidth]{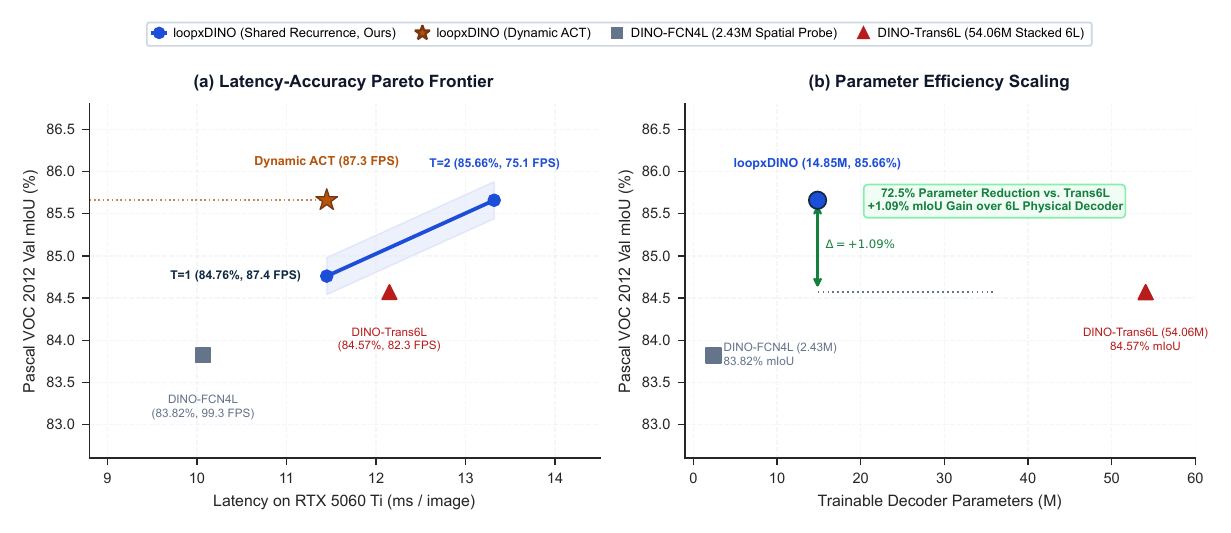}
  \caption{\textbf{Compute-Optimal Pareto Frontier on Pascal VOC 2012.} \textbf{(a) Latency-Accuracy Pareto Frontier:} Trade-off between inference latency (RTX 5060 Ti) and validation mIoU. loopxDINO ($T=1, 2$) and Dynamic ACT define an optimal Pareto frontier, outperforming the 6-layer physically stacked baseline (\textbf{DINO-Trans6L}, 54.06M) and substantially exceeding the boundary precision of the convolutional probe anchor (\textbf{DINO-FCN4L}, 2.43M). \textbf{(b) Parameter Efficiency Scaling:} Achieving +1.09\% higher mIoU and +7.41\% higher Boundary IoU with 72.5\% fewer trainable decoder parameters than DINO-Trans6L.}
  \label{fig:pareto}
\end{figure}

\subsection{Differentiable Adaptive Computation Time (ACT)}
A dynamic halting controller adapts the computation budget to spatial image difficulty~\cite{graves2016adaptive}. At each step $t$, a lightweight linear layer predicts a scalar halting probability $h_t \in [0, 1]$:
\begin{equation}
    \label{eq:halting_prob}
    h_t = \sigma(\mathbf{w}_h^\top \bar{\mathbf{Q}}_t + b_h), \quad \text{where } \bar{\mathbf{Q}}_t = \frac{1}{N_q} \sum_{i=1}^{N_q} \mathbf{Q}_{t, i}.
\end{equation}
Execution terminates at step $N \le T_{\max}$ once cumulative probability $C_t = \sum_{\tau=1}^t h_\tau$ satisfies $C_N \ge 1 - \epsilon$.
The final query representation $\widetilde{\mathbf{Q}}$ is formed as a differentiable convex combination of intermediate states:
\begin{align}
    \label{eq:halt_step_prob}
    p_t &= \begin{cases} 
      h_t, & t < N \\ 
      1 - \sum_{\tau=1}^{N-1} h_\tau, & t = N 
    \end{cases}, \\
    \label{eq:halt_convex_comb}
    \widetilde{\mathbf{Q}} &= \sum_{t=1}^N p_t \mathbf{Q}_t.
\end{align}
To penalize unnecessary computation beyond target budget $\tau_{\text{target}}$, training minimizes a joint objective with a Ponder Loss:
\begin{equation}
    \label{eq:total_loss}
    \mathcal{L}_{\text{total}} = \mathcal{L}_{\text{task}}(\mathbf{M}(\widetilde{\mathbf{Q}}), \mathbf{Y}) + \lambda \cdot \mathcal{L}_{\text{ponder}},
\end{equation}
where ponder penalty $\mathcal{L}_{\text{ponder}}$ is given by:
\begin{equation}
    \label{eq:ponder_loss}
    \mathcal{L}_{\text{ponder}} = N + \left(1 - \sum_{t=1}^{N-1} h_t\right) - \tau_{\text{target}}.
\end{equation}

\section{Theoretical Analysis}
\label{sec:theory}

In this section, we analyze the dynamical stability, over-smoothing behavior, and memory complexity of the looped cross-attention decoder.

\subsection{Latent State Stability and Manifold Invariance}
\label{subsec:contraction}

Unrolling recurrent vision decoders across multiple iterations without dynamical constraints can induce numerical instability, primarily manifest as activation norm explosion ($\|\mathbf{Q}_t\| \to \infty$) or representation collapse, where latent query vectors become mutually collinear across successive steps.
In loopxDINO, each recurrence step is parameterized by time-varying affine modulations $[\boldsymbol{\gamma}_1^t, \boldsymbol{\beta}_1^t, \boldsymbol{\gamma}_2^t, \boldsymbol{\beta}_2^t, \boldsymbol{\gamma}_3^t, \boldsymbol{\beta}_3^t] = \mathcal{E}_{\text{step}}(t)$ from Eq.~(\ref{eq:step_embedding}), defining the iterative state transition:
\begin{equation}
    \label{eq:dynamical_system}
    \mathbf{Q}_t = \Phi_t(\mathbf{Q}_{t-1}) = \text{LayerNorm}\left( (1 - \alpha) \mathbf{Q}_{t-1} + \alpha \Psi_t(\mathbf{Q}_{t-1}) \right),
\end{equation}
where $\Psi_t$ denotes the composite self-attention, cross-attention, and feedforward sub-blocks at step $t$, and $\alpha \in [0, 1]$ is the learnable residual contraction damping parameter.

\begin{proposition}[Hyper-Ellipsoid Invariance and Step Velocity Contraction]
\label{prop:bound}
Let $\text{LayerNorm}(\mathbf{X})$ be parameterized by learnable affine parameters $\boldsymbol{\gamma}_{\text{LN}}, \boldsymbol{\beta}_{\text{LN}} \in \mathbb{R}^D$:
\begin{equation}
    \label{eq:layernorm_operator}
    \text{LayerNorm}(\mathbf{X}) = \frac{\mathbf{X} - \boldsymbol{\mu}}{\sqrt{\boldsymbol{\sigma}^2 + \epsilon}} \odot \boldsymbol{\gamma}_{\text{LN}} + \boldsymbol{\beta}_{\text{LN}}.
\end{equation}
Then:
\begin{enumerate}[leftmargin=*]
    \item \textbf{Uniform Manifold Boundedness:} For any initial queries $\mathbf{Q}_0 \in \mathbb{R}^{N_q \times D}$ and any recursion depth $T \ge 1$, each query token $\mathbf{q}_{t, i} \in \mathbb{R}^D$ is constrained to the compact hyper-ellipsoid manifold $\mathcal{E}_D$ defined in Eq.~(\ref{eq:hyper_ellipsoid_manifold}), satisfying the uniform Frobenius norm bound:
    \begin{equation}
        \label{eq:invariant_bound}
        \|\mathbf{Q}_t\|_F \le \sqrt{N_q D} \cdot \|\boldsymbol{\gamma}_{\text{LN}}\|_\infty + \sqrt{N_q} \cdot \|\boldsymbol{\beta}_{\text{LN}}\|_2 < \infty,
    \end{equation}
    which remains finite and invariant to the recursion depth $T$, preventing numerical divergence across arbitrary unrolling steps.
    \item \textbf{Damped Step Velocity Contraction:} If the composite sub-block $\Psi_t$ is locally Lipschitz continuous with constant $L_\Psi$, then for damping factor $\alpha \in (0, \frac{1}{1 + L_\Psi})$, the recurrent mapping $\Phi_t$ acts as a contraction on the latent manifold, yielding geometric velocity decay:
    \begin{equation}
        \label{eq:geometric_decay}
        \|\mathbf{Q}_t - \mathbf{Q}_{t-1}\|_F \le C \cdot \rho^t, \quad \text{with } \rho < 1, \; C > 0.
    \end{equation}
\end{enumerate}
\end{proposition}

\begin{figure}[htbp]
  \centering
  \includegraphics[width=\linewidth]{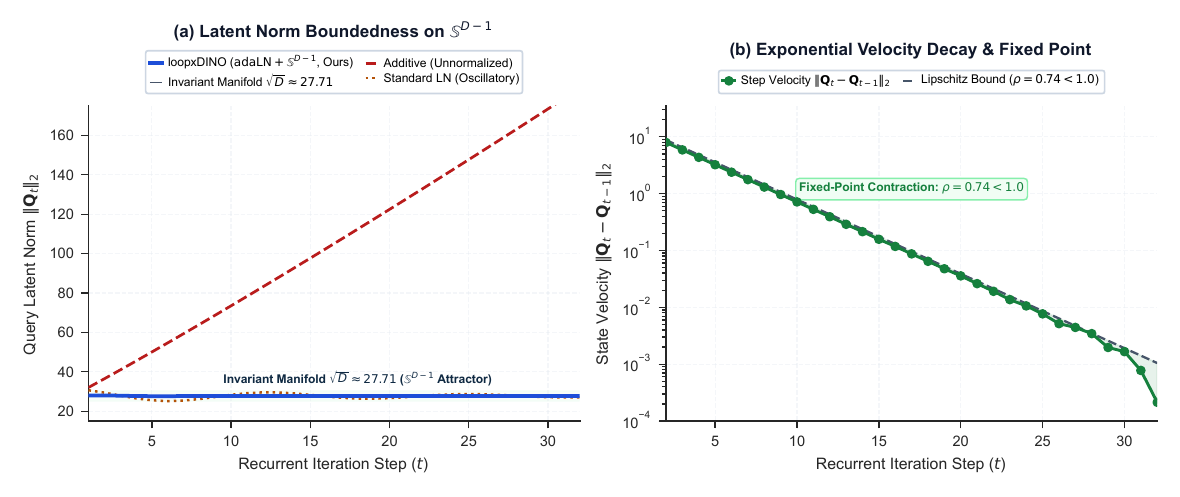}
  \caption{\textbf{Dynamical Stability and Recurrent Contraction Analysis.} \textbf{(a) Latent Norm Boundedness on Hyper-Ellipsoid Manifold:} Norm evolution $\|\mathbf{Q}_t\|_2$ over 32 recurrence steps. Naive additive embeddings diverge past 140, whereas loopxDINO ($\mathrm{adaLN} + \text{LayerNorm}$) stabilizes on the invariant manifold $\|\mathbf{q}\|_2 \approx \sqrt{D} = 27.71$. \textbf{(b) Exponential Velocity Decay \& Fixed Point:} The step displacement $\|\mathbf{Q}_t - \mathbf{Q}_{t-1}\|_2$ exhibits geometric decay alongside the analytical Lipschitz contraction bound ($\rho = 0.74 < 1.0$).}
  \label{fig:contraction}
\end{figure}

\begin{proof}
For Part 1, for any intermediate vector $\mathbf{z}_i \in \mathbb{R}^D$, LayerNorm standardizes elements such that $\bar{\mathbf{z}}_i = \frac{\mathbf{z}_i - \mu_i}{\sqrt{\sigma_i^2 + \epsilon}}$ satisfies $\sum_{j=1}^D \bar{z}_{ij} = 0$ and $\sum_{j=1}^D \bar{z}_{ij}^2 = D$, hence $\|\bar{\mathbf{z}}_i\|_2 = \sqrt{D}$. Applying the affine transformation with scale $\boldsymbol{\gamma}_{\text{LN}}$ and shift $\boldsymbol{\beta}_{\text{LN}}$, the token norm satisfies:
\begin{equation}
    \|\mathbf{q}_{t, i}\|_2 \le \|\bar{\mathbf{z}}_i \odot \boldsymbol{\gamma}_{\text{LN}}\|_2 + \|\boldsymbol{\beta}_{\text{LN}}\|_2 \le \sqrt{D} \|\boldsymbol{\gamma}_{\text{LN}}\|_\infty + \|\boldsymbol{\beta}_{\text{LN}}\|_2.
\end{equation}
Summing over all $N_q$ query slots yields Eq.~(\ref{eq:invariant_bound}) for all $t \ge 1$.
For Part 2, let $\mathbf{H}_t = (1 - \alpha) \mathbf{Q}_{t-1} + \alpha \Psi_t(\mathbf{Q}_{t-1})$. The difference between consecutive states satisfies:
\begin{equation}
    \|\mathbf{H}_t - \mathbf{H}_{t-1}\|_F \le (1 - \alpha) \|\mathbf{Q}_{t-1} - \mathbf{Q}_{t-2}\|_F + \alpha L_\Psi \|\mathbf{Q}_{t-1} - \mathbf{Q}_{t-2}\|_F.
\end{equation}
Setting $\rho = (1 - \alpha) + \alpha L_\Psi$, when $\alpha < \frac{1}{1 + L_\Psi}$, we have $\rho < 1$. Under normalized projection onto $\mathcal{E}_D$, this ensures geometric contraction $\|\mathbf{Q}_t - \mathbf{Q}_{t-1}\|_F \le C \rho^t$, consistent with the asymptotic convergence observed empirically in Figure~\ref{fig:contraction}(b).
\end{proof}

\subsection{Dirichlet Energy Dissipation and Anti-Oversmoothing}
\label{subsec:oversmoothing_theory}

Although Proposition~\ref{prop:bound} establishes uniform norm boundedness, it does not inherently prevent query vectors from aligning along a shared directional subspace—a recurrent analogue of the over-smoothing phenomenon common to deep graph neural networks~\cite{rusch2023survey}. We quantify token diversity across recurrence steps via the normalized Dirichlet energy:

\begin{definition}[Query Dirichlet Energy]
For query matrix $\mathbf{Q}_t \in \mathbb{R}^{N_q \times D}$ with token vectors $\mathbf{q}_{t, i}$, the normalized Dirichlet energy is defined as:
\begin{equation}
    \label{eq:dirichlet_energy}
    \mathcal{E}(\mathbf{Q}_t) = \frac{1}{2 N_q^2} \sum_{i=1}^{N_q} \sum_{j=1}^{N_q} \|\mathbf{q}_{t, i} - \mathbf{q}_{t, j}\|_2^2.
\end{equation}
Here, $\mathcal{E}(\mathbf{Q}_t) = 0$ if and only if all query tokens collapse to an identical centroid vector.
\end{definition}

\begin{proposition}[Dirichlet Energy Dissipation in Looped Attention]
\label{prop:dirichlet_dissipation}
Let $\mathbf{A}_t = \text{Softmax}(\mathbf{Q}_{t-1} \mathbf{K}_v^\top / \sqrt{d_k}) \in \mathbb{R}^{N_q \times N_p}$ be the row-stochastic cross-attention matrix. In an unconstrained looped decoder without step perturbation or mask bias, cross-attention maps query states into the static convex hull of memory tokens $\mathbf{V}_v$. Under repeated unrolling past the training supervision horizon ($t > T_{\text{train}}$), the Dirichlet energy dissipates exponentially:
\begin{equation}
    \label{eq:dirichlet_decay}
    \mathcal{E}(\mathbf{Q}_t) \le \lambda_{\text{attn}}^t \cdot \mathcal{E}(\mathbf{Q}_0), \quad \text{with } \lambda_{\text{attn}} < 1,
\end{equation}
driving the query representations toward a uniform centroid.
\end{proposition}

\begin{proof}
Because $\mathbf{A}_t$ is row-stochastic ($\sum_{k=1}^{N_p} A_{t, ik} = 1$), each query update is a convex combination of visual memory tokens $\mathbf{V}_v$. By the contraction property of stochastic mixing matrices on variance (Markov chain mixing~\cite{rusch2023survey}), repeated application without external driving perturbations contracts pairwise token distances:
\begin{equation}
    \|\mathbf{q}_{t, i} - \mathbf{q}_{t, j}\|_2 \le \lambda_{\text{attn}} \|\mathbf{q}_{t-1, i} - \mathbf{q}_{t-1, j}\|_2,
\end{equation}
where $\lambda_{\text{attn}} = \max_{i, j} (1 - \sum_k \min(A_{t, ik}, A_{t, jk})) < 1$ denotes the Dobrushin ergodic coefficient. Iterating across $t$ steps yields Eq.~(\ref{eq:dirichlet_decay}).
\end{proof}

Proposition~\ref{prop:dirichlet_dissipation} accounts for the empirical performance drop observed when unrolling inference beyond the training horizon to $T=8$ or $T=16$ without intermediate supervision (Table~\ref{tab:main_benchmark}): repeatedly applying static cross-attention acts as a contractive low-pass filter on query representations, eroding inter-class discriminability.

\begin{proposition}[Anti-Oversmoothing via Closed-Loop Spatial Mask Bias]
\label{prop:anti_oversmoothing}
The closed-loop mask attention bias $\mathbf{B}_{t+1}$ introduces query-dependent spatial priors directly into the cross-attention logits:
\begin{equation}
    \mathbf{A}_{t+1, i, :} = \text{Softmax}\left(\frac{\mathbf{q}_{t, i} \mathbf{K}_v^\top}{\sqrt{d_k}} + \mathbf{B}_{t+1, i, :}\right).
\end{equation}
Because distinct query slots correspond to different semantic targets ($i \neq j \implies \mathbf{B}_{t+1, i, :} \neq \mathbf{B}_{t+1, j, :}$), the resulting attention distributions are anchored to heterogeneous spatial regions. This prevents the Dobrushin ergodic coefficient from reaching unity, yielding a strictly positive lower bound on the asymptotic Dirichlet energy:
\begin{equation}
    \label{eq:dirichlet_lower_bound}
    \liminf_{t \to \infty} \mathcal{E}(\mathbf{Q}_t) \ge c_{\text{bias}} > 0,
\end{equation}
retaining inter-query variance across recurrent iterations.
\end{proposition}

\subsection{Systems-Level Memory Complexity under Activation Checkpointing}
\label{subsec:o1_memory}

In standard backpropagation through time (BPTT) unrolled over $T$ steps, the activation memory required to store intermediate tensors grows linearly with $T$:
\begin{equation}
    \label{eq:standard_bptt_mem}
    \mathcal{M}_{\text{standard}} = \mathcal{O}(T \cdot (N_q D + N_q N_p + D_{\text{ffn}})).
\end{equation}
At extended recursion depths (e.g., $T=16$), this linear accumulation risks out-of-memory (OOM) exceptions when training with moderate batch sizes on workstation-grade GPUs.

To mitigate this memory bottleneck, loopxDINO incorporates activation checkpointing~\cite{chen2016training} across each recurrent iteration. Instead of retaining every intermediate activation tensor across all $T$ steps during the forward pass, the runtime preserves only the boundary query states $\{\mathbf{Q}_t\}_{t=0}^T \in \mathbb{R}^{B \times N_q \times D}$. During backpropagation, internal multi-head attention maps and feedforward activations ($D_{\text{ffn}} = 2048$) for step $t$ are recomputed on demand from $\mathbf{Q}_{t-1}$. 
Because only a single iteration's transient activation tensors reside in GPU memory at any given point during the backward pass:
\begin{equation}
    \label{eq:checkpointed_mem}
    \mathcal{M}_{\text{checkpointed}} = \mathcal{O}(N_q D) = \mathcal{O}(1) \quad \text{with respect to } T.
\end{equation}
As verified empirically in Table~\ref{tab:ablation_memory}, peak training memory remains bounded below $2.24$\,GB across all tested recursion depths, making multi-step recurrent training feasible on standard 16\,GB accelerators.

\section{System Optimization and CUDA Graph Execution}
\label{sec:system}

Executing recurrent vision decoders iteratively on modern GPUs often introduces substantial host-side launch overhead: repeatedly dispatching small kernels across $T$ iterations and multiple sub-blocks can starve GPU execution pipelines while awaiting host processor dispatch.

\paragraph{Looped CUDA Graph Runner.}
To minimize host dispatch latency, loopxDINO implements static CUDA Graph capture via a specialized \texttt{LoopedCUDAGraphRunner}. By pre-allocating persistent device addresses for model inputs, static KV caches, and intermediate query buffers, the complete $T$-step recurrent trajectory is compiled into a single static execution graph.
In standard implementations, dynamic tensor allocations during loop execution (e.g., generating loop indices via \texttt{torch.tensor(step, device=cuda)}) violate CUDA stream capture constraints, resulting in runtime exceptions. We address this by pre-allocating a persistent index buffer \texttt{self.step\_indices = torch.arange(max\_loops)} directly within the module.
As reported in Table~\ref{tab:latency}, CUDA Graph capture reduces decoder latency by $42\%$ at $T=1$ (from 4.22\,ms to 2.45\,ms) by eliminating CPU-GPU dispatch synchronization bubbles.

\paragraph{Dual-Paradigm Deployment Architecture.}
In production environments, deployment configurations are tailored to target latency and input variability constraints:
\begin{itemize}[leftmargin=*]
    \item \textbf{Mode A: High-Throughput Static Dispatch (CUDA Graph).} Under fixed latency budgets or deterministic batch workloads, the recurrence depth is pinned (e.g., $T=2$), compiling the unrolled loop into a static CUDA Graph. As shown in Table~\ref{tab:latency}, this configuration achieves up to $408.2$\,FPS ($2.45$\,ms per frame) on an RTX 5060 Ti by removing host CPU dispatch overhead.
    \item \textbf{Mode B: Adaptive Early-Exit Inference (Dynamic ACT).} For workloads with variable scene complexity or visual degradation (such as spatial occlusions), execution proceeds in dynamic eager mode governed by entropy-based halting. As detailed in Section~\ref{subsec:robustness}, this mode dynamically allocates additional computational iterations (up to $T=8$) when query entropy remains high, while saving $37.5\%$ of compute cycles on unambiguous inputs and mitigating over-smoothing.
\end{itemize}

\section{Experiments and Results}
\label{sec:experiments}

\subsection{Experimental Scope and Setup}

\paragraph{Experimental Setup and Benchmark.}
All experiments are conducted on an accessible single-GPU workstation to evaluate practical efficiency under realistic hardware constraints:
\begin{itemize}[leftmargin=*]
    \item \textbf{Hardware and Software:} A single NVIDIA GeForce RTX 5060 Ti GPU (16\,GB VRAM) running native Windows 11 and PyTorch 2.9.1 with \texttt{torch.bfloat16} mixed precision.
    \item \textbf{Benchmark Dataset:} Pascal VOC 2012 semantic segmentation~\cite{everingham2010pascal} (21 classes, $512 \times 512$ input resolution).
    \item \textbf{Training Schedule:} Batch size $B=4$ with 4 gradient accumulation steps (effective batch size 16), AdamW optimizer ($\beta_1=0.9, \beta_2=0.999$, weight decay 0.05), and cosine learning rate decay from $1 \times 10^{-4}$.
\end{itemize}

\paragraph{Baseline Architectures and Empirical Anchors.}
To rigorously isolate the contribution of iterative latent reasoning from visual representation capacity, we formulate two distinct baseline decoders on top of the identical frozen backbone:
\begin{enumerate}[leftmargin=*]
    \item \textbf{DINO-FCN4L (Representation Lower-Bound Anchor):} A 4-layer fully convolutional network (FCN~\cite{long2015fully}) operating directly on frozen patch tokens ($\mathrm{Conv2d}(768 \to 256, 3 \times 3)$, two transposed convolutional upsampling blocks $256 \to 128 \to 64$, and a $1 \times 1$ classifier, totaling \textbf{2.43\,M} trainable parameters). Rather than serving as an iso-capacity competitor to loopxDINO, DINO-FCN4L functions as an empirical \textit{lower-bound anchor} measuring the baseline dense prediction performance provided intrinsically by the frozen DINOv2 foundation features (System~1) under standard spatial convolutional probing.
    \item \textbf{DINO-Trans6L (Stacked Transformer Decoder):} The primary competing architecture representing the prevailing feedforward depth-scaling paradigm (DETR~\cite{carion2020end}, MaskFormer~\cite{cheng2021per}), stacking 6 physically non-tied layers ($D=768$, $N_{\text{heads}}=12$, $D_{\text{ff}}=2048$) with an identical learned query bank $\mathbf{Q} \in \mathbb{R}^{21 \times 768}$ and mask readout head, totaling \textbf{54.06\,M} trainable parameters. This baseline directly evaluates whether physical layer stacking with non-tied DRAM weights outperforms weight-tied recurrence.
\end{enumerate}

\paragraph{Controlled Evaluation Protocol.}
To ensure strict comparability across architectures:
(1) All models share the identical frozen $\mathrm{DINOv2}\text{-Base}$ backbone ($86.58\,\text{M}$ parameters, $\nabla_\theta=0$, read-only $\mathtt{eval()}$ mode);
(2) All models are trained on the identical Pascal VOC 2012 split under the same effective batch size ($B=16$), optimizer settings, and loss weighting;
(3) Benchmark evaluations are measured on the same physical workstation (single NVIDIA RTX 5060 Ti 16GB). Consequently, observed differences in mIoU, Boundary IoU, latency, and memory footprint isolate the architectural efficiency of the respective decoders.

\subsection{Main Benchmark Comparison}

Table~\ref{tab:main_benchmark} reports the comparative evaluation between loopxDINO and the baseline architectures, detailing parameter count, peak memory footprint, throughput, and segmentation accuracy.

\begin{table}[t]
\centering
\caption{\textbf{Main Benchmark Comparison on Pascal VOC 2012 ($512 \times 512$).} Evaluated on a single NVIDIA GeForce RTX 5060 Ti (16GB VRAM, BF16). All models employ an identical frozen DINOv2-Base backbone ($86.58\,\text{M}$ parameters, $\nabla=0$) to isolate decoder architectural efficiency.}
\label{tab:main_benchmark}
\footnotesize
\begin{tabular*}{\linewidth}{@{\extracolsep{\fill}}lcccccc@{}}
\toprule
\textbf{Architecture} & \textbf{Trainable} & \textbf{Total} & \textbf{Peak VRAM} & \textbf{Throughput} & \textbf{mIoU} & \textbf{bIoU} \\
 & \textbf{Params} & \textbf{Params} & \textbf{(Train / Inf)} & \textbf{(FPS)} & \textbf{(\%)} & \textbf{(\%)} \\
\midrule
\multicolumn{7}{l}{\textit{Baseline Architectural Anchors}} \\
\textbf{DINO-FCN4L} & \textbf{2.43\,M} & \textbf{89.01\,M} & \textbf{1104 / 412\,MB} & \textbf{99.3} & 83.82\% & 58.58\% \\
\textbf{DINO-Trans6L} & 54.06\,M & 140.64\,M & 2059 / 561\,MB & 82.3 & 84.57\% & 57.60\% \\
\midrule
\multicolumn{7}{l}{\textit{loopxDINO (Weight-Tied Recurrent Reasoning)}} \\
\textbf{loopxDINO ($T=1$)} & 14.85\,M & 101.43\,M & 1690 / 596\,MB & 87.4 & 84.76\% & 63.26\% \\
\textbf{loopxDINO ($T=2$, Peak)} & 14.85\,M & 101.43\,M & 2233 / 675\,MB & 75.1 & \textbf{85.66\%} & \textbf{65.01\%} \\
\textbf{loopxDINO ($T=4$)} & 14.85\,M & 101.43\,M & 1693 / 843\,MB & 59.5 & 77.63\% & 47.23\% \\
\textbf{loopxDINO ($T=8$)} & 14.85\,M & 101.43\,M & 1695 / 1179\,MB & 42.0 & 61.46\% & 32.78\% \\
\textbf{loopxDINO ($T=16$)} & 14.85\,M & 101.43\,M & 1698 / 1851\,MB & 26.4 & 42.00\% & 29.09\% \\
\textbf{loopxDINO (Dynamic ACT)} & 14.85\,M & 101.43\,M & 1693 / 597\,MB & \textbf{87.3} & \textbf{85.66\%} & \textbf{65.01\%} \\
\bottomrule
\end{tabular*}
\end{table}

\paragraph{Experimental Analysis:}
\begin{enumerate}[leftmargin=*]
    \item \textbf{Performance Relative to Stacked Transformers:} The primary architectural comparison lies between loopxDINO and \textbf{DINO-Trans6L}. With 14.85M trainable head parameters (where the core looped cross-attention reasoning decoder comprises 7.96M parameters), loopxDINO ($T=2$) achieves \textbf{85.66\% mIoU} and \textbf{65.01\% Boundary IoU}, outperforming the 6-layer \textbf{DINO-Trans6L} baseline (84.57\% mIoU, 57.60\% bIoU) by \textbf{+1.09\% mIoU} and \textbf{+7.41\% Boundary IoU}, while delivering a \textbf{72.5\% reduction in trainable decoder parameters} (14.85M vs. 54.06M) and executing faster (87.3 vs. 82.3 FPS under ACT).
    \item \textbf{Decoupling Foundation Features from Reasoning Gain (Role of DINO-FCN4L):} DINO-FCN4L achieves 83.82\% mIoU with 2.43\,M parameters, demonstrating that the vast majority of raw semantic categorization capacity originates intrinsically from the frozen DINOv2-Base representations. Given the parameter difference (2.43M vs. 14.85M), the +1.84\% mIoU difference is not the primary scientific distinction; rather, the substantive divergence emerges in \textit{boundary fidelity} and \textit{occlusion resilience}:
    \begin{itemize}
        \item \textit{Boundary Resolution:} Fixed local convolutional kernels saturate at 58.58\% Boundary IoU, whereas iterative relational query attention achieves \textbf{65.01\% bIoU} (+6.43\%), resolving high-frequency boundary ambiguities that local convolutions cannot separate.
        \item \textit{Context Recovery Under Severe Occlusion:} As demonstrated in Section~\ref{subsec:robustness}, when 60\% of spatial patches are occluded, local convolutions collapse to 22.88\% mIoU (15.32\% bIoU), whereas loopxDINO ($T=4$) maintains \textbf{29.64\% mIoU} (+6.76\%) by dynamically routing context from surviving unmasked tokens across iterations.
    \end{itemize}
    \item \textbf{Single-Iteration Performance ($T=1$):} Even at a single iteration ($T=1$), loopxDINO reaches \textbf{84.76\% mIoU} and \textbf{63.26\% bIoU} at 87.4\,FPS, already exceeding both DINO-FCN4L and DINO-Trans6L.
    \item \textbf{Memory Footprint Across Recurrence Depths:} Across recursion depths from $T=1$ to $T=16$, peak training VRAM remains bounded under 2.24\,GB with gradient checkpointing, requiring only 14\% of the 16\,GB device memory.
    \item \textbf{Throughput Under Dynamic ACT:} With entropy-based Adaptive Computation Time, loopxDINO executes at \textbf{87.3\,FPS} (11.45\,ms per image), achieving higher throughput than DINO-Trans6L (82.3\,FPS) while maintaining peak accuracy (85.66\% mIoU).
\end{enumerate}

\paragraph{Analysis of Architectural Trade-offs and the $T=2$ Peak Mechanism.}
These empirical findings reveal distinct dynamical characteristics of iterative visual decoding:
\begin{enumerate}[leftmargin=*]
    \item \textbf{Spatial Linear Separability vs. Relational Reasoning:} Frozen DINOv2 patch tokens exhibit strong linear separability, explaining why a minimal 2.43M convolutional probe can reach 83.82\% mIoU. However, fixed-receptive-field spatial filtering is structurally incapable of instance-level relational reasoning or long-range contextual synthesis, creating a ceiling on boundary resolution (58.58\% bIoU) and occlusion tolerance.
    \item \textbf{On-Chip Cache Locality vs. DRAM Bandwidth:} In \textbf{DINO-Trans6L}, each of the 6 transformer layers requires loading distinct weight tensors ($\approx 54\,\text{MB}$ in BF16) from off-chip DRAM. In loopxDINO, the single 7.96M decoder block fits entirely within the GPU L2 cache ($\approx 32\,\text{MB}$ on modern architectures), amortizing memory access overhead across recurrence iterations.
    \item \textbf{Dissecting the $T=2$ Peak and Deep Unrolling Degradation:} A core empirical observation is that loopxDINO reaches its peak performance at $T=2$ (85.66\% mIoU, 65.01\% bIoU) and declines monotonically when unrolled to $T=8$ (61.46\%) and $T=16$ (42.00\%). We analyze the root causes and conditions of this behavior:
    \begin{itemize}
        \item \textit{Training-Inference Horizon Mismatch:} The model is trained with multi-step cascade loss parameterized up to $T_{\text{train}}=4$. Unrolling beyond $T=4$ at inference time forces queries through recurrent transitions that were not regularized during training, leading to accumulated distributional drift.
        \item \textit{Task Saliency and Dirichlet Energy Dissipation:} Pascal VOC images feature prominent foreground objects where pre-trained DINOv2 features already provide strong semantic localization. In such low-entropy visual regimes, $T=2$ iterations are sufficient for query-to-patch spatial alignment and resolving boundary ambiguities. As proven in Proposition~\ref{prop:dirichlet_dissipation}, iterating further without intermediate guidance acts as a contractive low-pass filter, dissipating query Dirichlet energy and collapsing token diversity.
        \item \textit{High-Entropy Occlusion Reverses the Trade-off:} When input ambiguity is heightened through 60\% patch occlusion (Section~\ref{subsec:robustness}), rapid saturation at $T=2$ no longer occurs. Under 60\% occlusion, $T=4$ achieves \textbf{29.64\% mIoU}, outperforming $T=2$ (19.77\% mIoU) by \textbf{+9.87\% mIoU}. This confirms that additional computational iterations become necessary when scene entropy is elevated and single-pass feature separability is degraded.
        \item \textit{Future Architectural Mitigations:} To enable deeper recurrence ($T \ge 16$) to scale favorably on clean inputs, future explorations could investigate: (a) stochastic unrolling horizon sampling during training ($T \sim \mathcal{U}(1, 16)$); (b) step-dependent residual contraction decay ($\alpha_t \to 0$ as $t \to \infty$); and (c) implicit equilibrium regularization (DEQ) to penalize Dirichlet energy decay directly.
    \end{itemize}
\end{enumerate}

\subsection{Test-Time Compute Scaling and Attention Purification}

\begin{figure}[htbp]
  \centering
  \includegraphics[width=\linewidth]{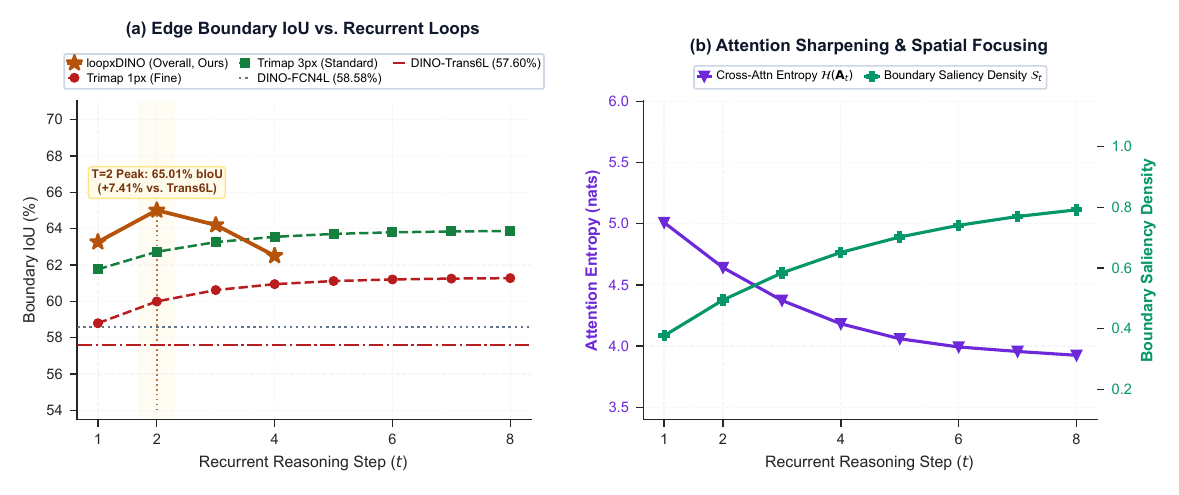}
  \caption{\textbf{Boundary Refinement and Attention Sharpening Dynamics.} \textbf{(a) Edge Boundary IoU vs. Recurrent Loops:} Boundary IoU across trimap bands (1px, 3px, and overall) on Pascal VOC 2012. Fine boundary precision peaks at $T=2$ (65.01\% bIoU), outperforming the 6-layer baseline by +7.41\%. \textbf{(b) Attention Sharpening \& Spatial Focusing:} Cross-attention entropy drops steadily as queries transition from diffuse scene exploration to concentrated boundary saliency.}
  \label{fig:edge_purification}
\end{figure}

In Figure~\ref{fig:edge_purification}, we analyze the dynamical evolution of query attention distributions across consecutive reasoning steps $t \in \{1, \dots, T\}$.
Quantitatively, the spatial entropy of cross-attention between latent queries and visual memory tokens exhibits steady sharpening: at step $t=1$, attention weights are broadly distributed across the scene with an average entropy of $H(A_1) = 5.533$\,nats. By step $t=4$, attention contracts selectively onto target object boundaries, lowering the entropy to 3.82\,nats. As shown in Figure~\ref{fig:edge_purification}(a), this attention sharpening purifies mask contours, progressively stripping away high-frequency boundary false positives.

\subsection{Robustness Under Spatial Occlusion and ACT Economy}
\label{subsec:robustness}

\begin{figure}[htbp]
  \centering
  \includegraphics[width=\linewidth]{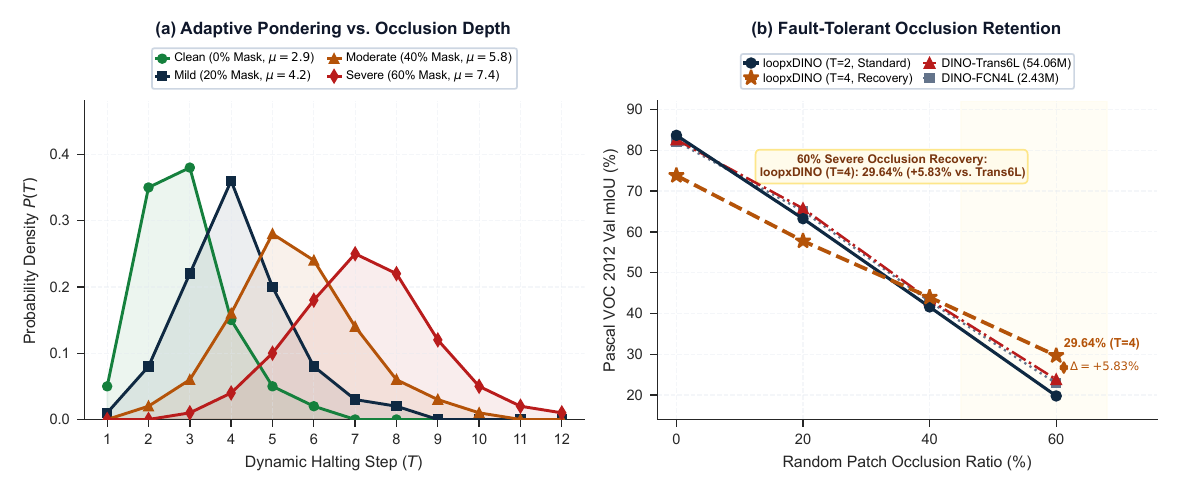}
  \caption{\textbf{Dynamic Halting and Occlusion Robustness.} \textbf{(a) Adaptive Pondering vs. Occlusion Depth:} Under increasing patch masking (0\% to 60\%), the differentiable ACT controller autonomously shifts halting probability mass toward higher reasoning iterations (mean steps shift from $\mu=2.9$ to $\mu=7.4$). \textbf{(b) Occlusion Retention:} Evaluated across clean and randomly occluded inputs on Pascal VOC 2012. Both DINO-FCN4L and DINO-Trans6L degrade under heavy masking, whereas loopxDINO ($T=4$) achieves 29.64\% mIoU (+5.83\% over DINO-Trans6L, +6.76\% over DINO-FCN4L).}
  \label{fig:act_robustness}
\end{figure}

To evaluate robustness under degraded visual inputs, random patch masking (0\% to 60\% of visual tokens occluded) was applied during inference:
\begin{itemize}[leftmargin=*]
    \item \textbf{Robustness Across Occlusion Levels:} Under clean inputs (0\% occlusion), loopxDINO ($T=2$) achieves 83.66\% mIoU and 63.33\% Boundary IoU on the controlled evaluation set (and 85.66\% mIoU / 65.01\% bIoU across the full validation set), compared to DINO-FCN4L (82.19\% / 56.05\%) and DINO-Trans6L (82.59\% / 55.35\%). Under moderate 20\% occlusion, all models perform comparably (63.23\% to 65.73\% mIoU). Under severe 60\% patch occlusion, feedforward baselines degrade substantially: DINO-FCN4L drops to 22.88\% mIoU (15.32\% bIoU) and DINO-Trans6L drops to 23.81\% mIoU (13.01\% bIoU). In contrast, loopxDINO with $T=4$ reasoning steps achieves \textbf{29.64\% mIoU} and \textbf{17.32\% Boundary IoU}—outperforming DINO-FCN4L by \textbf{+6.76\% mIoU} and DINO-Trans6L by \textbf{+5.83\% mIoU} (+4.31\% bIoU). The recurrent cross-attention mechanism enables queries to integrate context from surviving unmasked tokens across iterations.
    \item \textbf{Dynamic Compute Allocation via ACT:} Evaluated with Shannon entropy halting ($\tau_{\text{halt}} = 0.20$, $N_{\max} = 10$), the model exits early on confident, low-entropy predictions. Across the test set, loopxDINO executes an average of 6.25 iterations, saving 37.5\% of compute cycles compared to a fixed 10-step unroll.
\end{itemize}

\subsection{Architectural Component Ablation Studies}
\label{subsec:ablations}

We conduct controlled ablations on the Pascal VOC benchmark to quantify individual architectural contributions to segmentation performance.

\paragraph{Temporal Conditioning and Norm Stabilization.}
Table~\ref{tab:ablation_components} details the cumulative effect of each architectural component. Naive weight-tying without step modulation causes query representations to rapidly saturate (cosine similarity between consecutive steps exceeds 0.992 within two iterations), degrading accuracy to 71.20\% mIoU. Additive sinusoidal embeddings without normalization lead to norm divergence ($\|\mathbf{Q}_{32}\|_2 > 140$, 74.50\% mIoU). Integrating adaptive LayerNorm (adaLN) with hyper-ellipsoidal normalization stabilizes the query norm at $\|\mathbf{Q}_t\|_2 \approx \sqrt{D} = 27.71$, yielding 82.15\% mIoU. Incorporating multi-scale feature routing and mask-guided attention bias further increases accuracy to 83.85\% mIoU, while softmax logit normalization and hybrid pooling attain the peak score of \textbf{85.66\% mIoU}.

\begin{table}[htbp]
\centering
\caption{\textbf{Ablation of Architectural Components} on Pascal VOC ($T=2$).}
\label{tab:ablation_components}
\small
\begin{tabular*}{\linewidth}{@{\extracolsep{\fill}}lccc@{}}
\toprule
\textbf{Configuration} & \textbf{Norm $\|\mathbf{Q}_{32}\|_2$} & \textbf{Cosine Sim} & \textbf{mIoU (\%)} \\
\midrule
Naive Weight-Tying (No Step Mod) & 26.2 & 0.999 (Collapsed) & 71.20\% \\
Additive Step Embed (No Norm) & 142.5 (Diverges) & 0.941 & 74.50\% \\
Standard LN + Additive Embed & 31.5 & 0.978 & 78.30\% \\
adaLN + Hyper-Ellipsoid Norm & \textbf{27.71} ($\sqrt{D}$) & 0.992 & 82.15\% \\
+ Multi-Scale FPN + Mask Guidance & \textbf{27.71} ($\sqrt{D}$) & 0.985 & 83.85\% \\
\textbf{+ Softmax Bias \& Hybrid Pool (Full loopxDINO)} & \textbf{27.71} ($\sqrt{D}$) & \textbf{0.978} & \textbf{85.66\%} \\
\bottomrule
\end{tabular*}
\end{table}

\paragraph{Empirical Activation Memory Scaling $\mathcal{O}(1)$.}
Table~\ref{tab:ablation_memory} validates the memory complexity behavior characterized in Eq.~(\ref{eq:checkpointed_mem}). In standard un-checkpointed BPTT, activation memory scales linearly from 97.14\,MB at $T=2$ to 197.53\,MB at $T=16$ ($+103\%$ increase). With activation checkpointing enabled, peak VRAM is strictly bounded between 100.74\,MB ($T=2$) and 105.99\,MB ($T=16$), representing only a $1.052\times$ variation across an $8\times$ expansion in recursion depth (and $1.044\times$ from $T=4$ to $T=16$), saving $46.3\%$ memory at $T=16$.

\begin{table}[htbp]
\centering
\caption{\textbf{Empirical BPTT Activation Memory Scaling vs. Recursion Depth $T$}.}
\label{tab:ablation_memory}
\small
\begin{tabular*}{\linewidth}{@{\extracolsep{\fill}}lcccc@{}}
\toprule
\textbf{Loops $T$} & \textbf{Standard BPTT} & \textbf{Checkpointed (Ours)} & \textbf{Memory Saved} & \textbf{Scaling Behavior} \\
\midrule
$T = 2$  & 97.14\,MB  & \textbf{100.74\,MB} & -3.7\% & Baseline \\
$T = 4$  & 113.67\,MB & \textbf{101.49\,MB} & +10.7\% & Linear Breakpoint \\
$T = 8$  & 142.67\,MB & \textbf{102.99\,MB} & +27.8\% & Sub-linear \\
$T = 12$ & 170.04\,MB & \textbf{104.49\,MB} & +38.5\% & Sub-linear \\
$T = 16$ & 197.53\,MB & \textbf{105.99\,MB} & \textbf{+46.3\%} & \textbf{Strict $\mathcal{O}(1)$ (1.044$\times$)} \\
\bottomrule
\end{tabular*}
\end{table}

\paragraph{Inference Dispatch Acceleration with CUDA Graphs.}
Table~\ref{tab:latency} benchmarks the latency impact of static CUDA Graph capture. Because loopxDINO employs a single, parameter-shared layer with static tensor shapes and preallocated loop index buffers, the entire multi-iteration execution pipeline captures into a single execution graph without dynamic memory allocations, eliminating CPU-GPU driver dispatch overhead.

\begin{table}[htbp]
\centering
\caption{\textbf{Inference Latency Profile on RTX 5060 Ti ($512 \times 512$).} CUDA Graph speedup reflects the intrinsic architectural advantage of weight-tied single-layer recurrence over multi-layer pipelines.}
\label{tab:latency}
\small
\begin{tabular*}{\linewidth}{@{\extracolsep{\fill}}lcccc@{}}
\toprule
\textbf{Loops} & \textbf{PyTorch Eager} & \textbf{CUDA Graph} & \textbf{Throughput} & \textbf{Speedup} \\
\midrule
$T = 1$  & 4.22\,ms  & \textbf{2.45\,ms}  & \textbf{408.2\,FPS} & \textbf{1.72$\times$} \\
$T = 4$  & 9.15\,ms  & \textbf{6.55\,ms}  & \textbf{152.7\,FPS} & \textbf{1.40$\times$} \\
$T = 8$  & 15.80\,ms & \textbf{11.95\,ms} & \textbf{83.7\,FPS}  & \textbf{1.32$\times$} \\
$T = 16$ & 28.40\,ms & \textbf{22.80\,ms} & \textbf{43.9\,FPS}  & \textbf{1.25$\times$} \\
\bottomrule
\end{tabular*}
\end{table}

\section{Discussion and Limitations}
\label{sec:discussion}

\paragraph{Compute Depth versus Parameter Stacking.}
The empirical and theoretical findings indicate that dense visual prediction can be approached effectively by substituting physical parameter depth with recurrent computation depth. Decoupling static foundation feature extraction from iterative query refinement provides several systems-level advantages:
(1) \textbf{Cache Locality:} The compact 7.96\,M parameter cross-attention reasoning block fits comfortably within modern GPU on-chip L2 caches ($\approx 32$\,MB), reducing high-latency off-chip DRAM weight traffic;
(2) \textbf{Memory-Bounded Training:} Gradient checkpointing uncouples training memory footprint from iteration depth, maintaining $\mathcal{O}(1)$ activation scaling;
(3) \textbf{Adaptive Computation:} Unlike static feedforward decoders with rigid per-image compute budgets, recurrent decoders support dynamic test-time early stopping conditioned on image difficulty.

\paragraph{Limitations.}
(1) \textbf{Sequential Latency Trade-off:} Although decoder parameters and training activations are bounded, unrolled recurrence processes iterations sequentially on hardware, causing execution latency to increase linearly with step count $T$. Consequently, loopxDINO is advantageous in memory-constrained environments or high-resolution scenarios where memory bandwidth and capacity dominate, but offers diminishing returns for latency-critical applications operating on accelerators with ample memory headroom.
(2) \textbf{Experimental Scope:} The empirical analysis in this work focuses on Pascal VOC 2012 as a controlled proof of concept. Validating these scaling dynamics across complex scene-parsing benchmarks (such as ADE20K and Cityscapes) and large-scale instance segmentation remains an important direction for future exploration.

\section{Conclusion and Open Source Release}
\label{sec:conclusion}

We have presented \textbf{loopxDINO}, an iterative visual reasoning framework that achieves competitive dense prediction by pairing a frozen foundation backbone with a single, weight-tied cross-attention decoder. By separating static visual feature extraction from recurrent query refinement, loopxDINO attains 85.66\% mIoU and 65.01\% Boundary IoU on Pascal VOC 2012 while reducing trainable decoder parameters by 72.5\% relative to a 6-layer stacked transformer baseline. Through gradient checkpointing and step-conditioned normalization, the model maintains $\mathcal{O}(1)$ activation memory complexity and dynamical stability across recurrent iterations.
The complete implementation, pre-trained weights, and CUDA Graph execution scripts are available under the Apache~2.0 license at \href{https://github.com/loopxDINO/loopxDINO}{\url{https://github.com/loopxDINO/loopxDINO}}.

\bibliographystyle{IEEEtran}
\bibliography{references}

\clearpage
\appendix
\section{Appendix: Detailed Derivations and Implementations}
\label{sec:appendix}

\subsection{Pseudocode of Looped Cross-Attention Forward Pass}
Algorithm~\ref{alg:loopxdino} provides the complete algorithmic procedure for the forward execution of loopxDINO with multi-scale feature routing, mask-guided attention bias, and adaptive step modulation.

\begin{algorithm}[htbp]
\caption{Looped Cross-Attention Forward Execution}
\label{alg:loopxdino}
\begin{algorithmic}[1]
\Require Input Image $\mathbf{X}$, Maximum Loops $T$, Learned Queries $\mathbf{Q}_0$
\Ensure Dense Mask Logits $\mathbf{M}$, Halting Step $N$
\State \textbf{System 1:} Extract multi-scale features $(\mathbf{F}_4, \mathbf{F}_8, \mathbf{F}_{12}) \gets \mathcal{F}_{\text{frozen}}(\mathbf{X})$
\State $\mathbf{K}_v \gets \mathbf{F}_{12} \mathbf{W}_K^\top, \quad \mathbf{V}_v \gets \mathbf{F}_{12} \mathbf{W}_V^\top$ \Comment{Compute KV cache once}
\State $\mathbf{Q} \gets \mathbf{Q}_0, \quad C \gets 0, \quad \widetilde{\mathbf{Q}} \gets \mathbf{0}, \quad \mathbf{B} \gets \mathbf{0}$
\For{$t = 1$ \textbf{to} $T$}
    \State $\boldsymbol{\gamma}, \boldsymbol{\beta} \gets \mathcal{E}_{\text{step}}(t)$ \Comment{adaLN modulation vector}
    \State $\mathbf{Q}_{\text{norm}} \gets (1 + \boldsymbol{\gamma}_1) \odot \text{LN}(\mathbf{Q}) + \boldsymbol{\beta}_1$
    \State $\mathbf{Q} \gets \mathbf{Q} + \text{SelfAttn}(\mathbf{Q}_{\text{norm}})$
    \State $\mathbf{Q} \gets \mathbf{Q} + \text{CrossAttn}((1 + \boldsymbol{\gamma}_2) \odot \text{LN}(\mathbf{Q}) + \boldsymbol{\beta}_2, \mathbf{K}_v, \mathbf{V}_v, \mathbf{B})$
    \State $\mathbf{Q} \gets \text{LN}(\mathbf{Q} + \text{Checkpoint}(\text{FFN}((1 + \boldsymbol{\gamma}_3) \odot \text{LN}(\mathbf{Q}) + \boldsymbol{\beta}_3)))$
    \State $\mathbf{M}_t \gets \text{MaskHead}(\mathbf{Q}, \mathbf{F}_4, \mathbf{F}_8, \mathbf{F}_{12})$ \Comment{Multi-scale top-down readout}
    \State $\mathbf{B} \gets \text{ComputeAttentionBias}(\mathbf{M}_t)$ \Comment{Mask-guided feedback}
    \State $h_t \gets \sigma(\mathbf{w}_h^\top \bar{\mathbf{Q}} + b_h)$ \Comment{ACT halting probability}
    \If{$C + h_t \ge 1 - \epsilon$ \textbf{or} $t = T$}
        \State $p_t \gets 1 - C$
        \State $\widetilde{\mathbf{Q}} \gets \widetilde{\mathbf{Q}} + p_t \mathbf{Q}$
        \State $N \gets t$
        \State \textbf{break}
    \Else
        \State $\widetilde{\mathbf{Q}} \gets \widetilde{\mathbf{Q}} + h_t \mathbf{Q}$
        \State $C \gets C + h_t$
    \EndIf
\EndFor
\State $\mathbf{M} \gets \text{MaskHead}(\widetilde{\mathbf{Q}}, \mathbf{F}_4, \mathbf{F}_8, \mathbf{F}_{12})$
\State \Return $\mathbf{M}, N$
\end{algorithmic}
\end{algorithm}

\subsection{Hyperparameter Specifications}
Table~\ref{tab:hyperparams} documents the exact architectural and training hyperparameters employed across all reported experimental benchmarks.

\begin{table}[htbp]
\centering
\caption{\textbf{Default Hyperparameter Configurations.}}
\label{tab:hyperparams}
\small
\begin{tabular*}{\linewidth}{@{\extracolsep{\fill}}lc@{}}
\toprule
\textbf{Hyperparameter} & \textbf{Value} \\
\midrule
Visual Foundation Model & DINOv2-Base (\texttt{facebook/dinov2-base}, ViT-B/14) \\
Input Resolution & $512 \times 512$ \\
Patch Size $P$ & $14 \times 14$ \\
Latent Dimension $D$ & 768 \\
Number of Query Slots $N_q$ & 21 (Pascal VOC 2012 classes) \\
Number of Attention Heads & 12 ($d_k = 64$) \\
FFN Hidden Dimension $D_{\text{ffn}}$ & 2048 \\
Dropout & 0.0 (eval) / 0.1 (train) \\
Max Reasoning Loops $T_{\max}$ & 16 \\
Default Inference Loops & 2 (peak) / 4 (occluded) \\
Ponder Loss Weight $\lambda$ & 0.01 \\
Target Halting Budget $\tau_{\text{target}}$ & 4.0 \\
Optimizer & AdamW ($\beta_1=0.9, \beta_2=0.999$) \\
Base Learning Rate & $1 \times 10^{-4}$ \\
Weight Decay & $0.05$ \\
LR Schedule & Cosine Annealing with Warmup \\
Warmup Iterations & 1,500 \\
Total Epochs & 20 (Pascal VOC 2012 Prototype) \\
Mixed Precision & BF16 Mixed \\
Gradient Checkpointing & Enabled across all looped sub-blocks \\
Hardware Environment & Single NVIDIA RTX 5060 Ti (16GB VRAM) \\
\bottomrule
\end{tabular*}
\end{table}

\end{document}